\documentclass[10pt,twocolumn,letterpaper]{article}

\usepackage[pagenumbers]{cvpr} 

\usepackage{graphicx}
\usepackage{amsmath}
\usepackage{amssymb}
\usepackage{booktabs}

\usepackage{times}
\usepackage{epsfig}
\usepackage{multirow}
\usepackage[ruled,vlined]{algorithm2e}
\usepackage[noend]{algpseudocode}
\usepackage{rotating}
\usepackage[table]{xcolor}
\usepackage{setspace}

\usepackage[pagebackref,breaklinks,colorlinks]{hyperref}

\usepackage[capitalize]{cleveref}
\crefname{section}{Sec.}{Secs.}
\Crefname{section}{Section}{Sections}
\Crefname{table}{Table}{Tables}
\crefname{table}{Tab.}{Tabs.}

\def\cvprPaperID{491} 
\def\confName{CVPR}
\def\confYear{2022}

\begin{document}

\title{DST: Dynamic Substitute Training for Data-free Black-box Attack}

\author{Wenxuan Wang \quad Xuelin Qian$^{\dagger}$\quad Yanwei Fu \quad Xiangyang Xue \\
Fudan University \\
\{wxwang19,xlqian,yanweifu,xyxue\}@fudan.edu.cn
}

\maketitle
\renewcommand{\thefootnote}%
{\fnsymbol{footnote}}
\footnotetext[2]{indicates corresponding author. }

\begin{abstract}
With the wide applications of deep neural network models in various computer vision tasks, more and more works study the model vulnerability to adversarial examples. For data-free black box attack scenario, existing methods are inspired by the knowledge distillation, and thus usually train a substitute model to learn knowledge from the target model using generated data as input. However, the substitute model always has a static network structure, which limits the attack ability for various target models and tasks. In this paper, we propose a novel dynamic substitute training attack method to encourage substitute model to learn better and faster from the target model. Specifically, a dynamic substitute structure learning strategy is proposed to adaptively generate optimal substitute model structure via a dynamic gate according to different target models and tasks. Moreover, we introduce a task-driven graph-based structure information learning constrain to improve the quality of generated training data, and facilitate the substitute model learning structural relationships from the target model multiple outputs. Extensive experiments have been conducted to verify the efficacy of the proposed attack method, which can achieve better performance compared with the state-of-the-art competitors on several datasets. 
\end{abstract}

\section{Introduction}

\begin{figure}[htbp]
\begin{centering}
\includegraphics[scale=0.4]{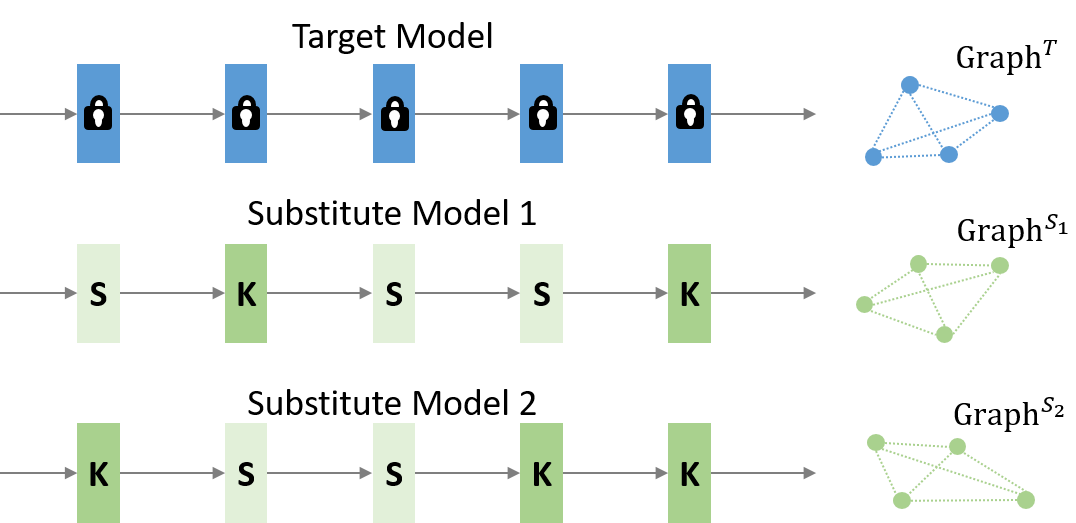}
\par\end{centering}
\caption{A conceptual overview of our method. Rather than retain all blocks, our approach learns to generate optimal substitute structure according to the black-box attack target. The lighter green blocks with `S' indicate the skipped blocks, and dark green blocks with `K' are the keeping blocks. The graph-based structural information is used to facilitate substitute training. 
\label{fig:intro}}
\end{figure}

Deep neural network models have achieved the state of the art performances in many challenging computer-vision tasks~\cite{he2016deep,simonyan2014very,krizhevsky2017imagenet}. These models have been wide-spread adopted in real-world applications, \emph{e.g.}, self-driving cars, license plate reading, disease diagnosis from medical images, and activity classification. However, recent studies \cite{szegedy2013intriguing,goodfellow2014explaining} show that deep neural networks are highly vulnerable to adversarial examples, which contain small and imperceptible perturbations crafted to fool the target models. This attracts more researchers to study the attack and defense for better assessing and improving the robustness of deep models.

Adversarial attack methods can be categorized into two main settings, \emph{i.e.}, white-box attack \cite{papernot2016limitations,carlini2017CW,kurakin2016adversarial,madry2017pgd,dong2018boost} and black-box attack \cite{chen2017zoo,ilyas2018prior,ilyas2018blackbox, dong2019efficient,chen2017zoo,cui2020substitute,gao2019boosting,yin2021adv}, by whether or not the attackers can have full access to the structure and parameters of the target model. Nowadays, in the era of big data, data is one of the most valuable assets for companies, and much of it also has privacy issues. Thus, in practical cases, it is not only difficult for attackers to know the details of the target model, but also hardly obtain the training data for the target model, even the number of categories. 

The purpose of this paper is to successfully achieve a data-free black-box attack according to the given target model. \textit{``Data-free''} suggests that we cannot access any knowledge about the data distribution (\emph{e.g.}, the type of data, number of categories, \emph{etc.}), which was used for the target model; \textit{``Black-box''} indicates the target model structure is completely shielded from the attackers, resulting in the limitation on getting model parameters or features of middle layers. The only available thing we can use is the output of probabilities/labels from the target model. Such a strict setting is more in line with the requirements of real-world scenarios, especially at a time when privacy data protection has attracted more attention. Inspired by the substitute training methods in black-box attack, many works \cite{zhou2020dast,wang2021delving} try to tackle the data-free black-attack through learning a substitute model for the target one with generated training data. However, there exist two main limitations in existing methods, (1) \emph{static substitute model structure for different targets:} due to the lack of prior knowledge, utilizing the same and static substitute model architecture for various target models or tasks definitely can not achieve powerful attack. Meanwhile, it is impractical and cost expensive to train multiple models to find the most suitable substitute model structure for each target variation. (2) \emph{Assumption of knowing the number of categories for the target model:} for the total data-free black-box attack, it is unreasonable for knowing the number of training data classes for the target model. Thus, using the label as the generator guidance information to synthesize diverse and label-controlled data for substitute training is unpractical.

In this study, to address the limitations of existing methods, we propose a novel and task-driven \textbf{D}ynamic \textbf{S}ubstitute \textbf{T}raining (DST) attack method for data-free black-box attacking, as illustrated in Fig.~\ref{fig:intro}.
Our DST attack adopts the basic substitute learning framework as in \cite{zhou2020dast,wang2021delving,kariyappa2021maze}, which generates the training data via a generator with noise as input, and takes advantage of the knowledge distillation concept to encourage the substitute model to has the same output as the target one when facing the same synthesized training image. 
To tackle the constant substitute model architecture problem (limitation (1)), in our DST attack algorithm, for the first time, we introduce a dynamic substitute structure learning strategy to automatically generate a more suitable substitute model structure according to different target models and tasks. To achieve such dynamic structure generation, we specially design a learnable dynamic gate to determine which blocks in the deep architecture can be skipped. 
To deal with being unaware of the prior knowledge about the training data classes issue (limitation (2)), we introduce a graph-based structural information learning strategy in DST, to further improve the generator performance and enhance the substitute training process. Such a learning strategy can facilitate the substitute model learning more implicit and detailed information from the structural relationship among multiple target model outputs. Meanwhile, such structural information can reflect the representation distance among a group of generated training data and stimulate the generator to deliver more valuable training data. 
Overall, our DST attack can adaptively generate optimal substitute model structure for various targets, improve the consistency between the substitute and target model, and encourage the generator to synthesize better training data via learning structural information. This can promote the data-free black-box attack performance.

The main contributions of this work are summarized below,
(1) We propose a novel and task-driven dynamic substitute training attack method to boost the data-free black-box attacking performance.
(2) For the first time, we introduce a dynamic substitute structure learning strategy to adaptively generate optimal substitute model architecture according to the different target models and tasks, instead of adopting the same and static network.
(3) To encourage the substitute model to learn more details from the target model and improve the quality of generated training data, we propose a graph-based structural information learning strategy to deeply explore the structural, relational, and valuable information from a bunch of target outputs.
(4) The comprehensive experiments over four public datasets and one online machine learning platform demonstrate that our DST method can achieve SOTA attack performance and significantly reduce the query times during the substitute training.

\section{Related Work}
\noindent \textbf{Adversarial Attack.}
Since deep learning models have achieved remarkable success on most computer vision tasks \cite{he2016deep,simonyan2014very,krizhevsky2017imagenet,qian2019leader,qian2020m,wang2020fm2u}, the study for the security of these models has attracted many researchers. \cite{szegedy2013intriguing} illustrates that deep neural networks are susceptible to adversarial perturbations. Subsequently, more and more works \cite{papernot2016limitations,carlini2017CW,kurakin2016adversarial,madry2017pgd,goodfellow2014explaining,dong2018boost,chen2017zoo,cui2020substitute,gao2019boosting,huang2019enhancing,yang2020learning,li2020towards} focus on the adversarial example generation task. In general, the attack task can be divided into white-box and black-box attacks, the former one can know the knowledge of the structure and parameters of the target model, and the latter one only has the access to the simple output of the target. Most white-box algorithms \cite{papernot2016limitations,carlini2017CW,kurakin2016adversarial,madry2017pgd,goodfellow2014explaining,dong2018boost} generate adversarial examples based on the gradient of loss function with respect to the inputs. For the black-box attack, some methods \cite{chen2017zoo,cui2020substitute,gao2019boosting} iteratively query the outputs of target model and estimate the gradient of target model via training a substitute model; and others \cite{huang2019enhancing,yang2020learning,li2020towards} focus on improving the transferability of adversarial examples across different models. In this work, we focus on the more practical and challenging scenario, \emph{i.e.}, the data-free black-box attack, which attacks the black-box target model without the need for any real data samples.

\begin{figure*}[htbp]
\begin{centering}
\includegraphics[scale=0.44]{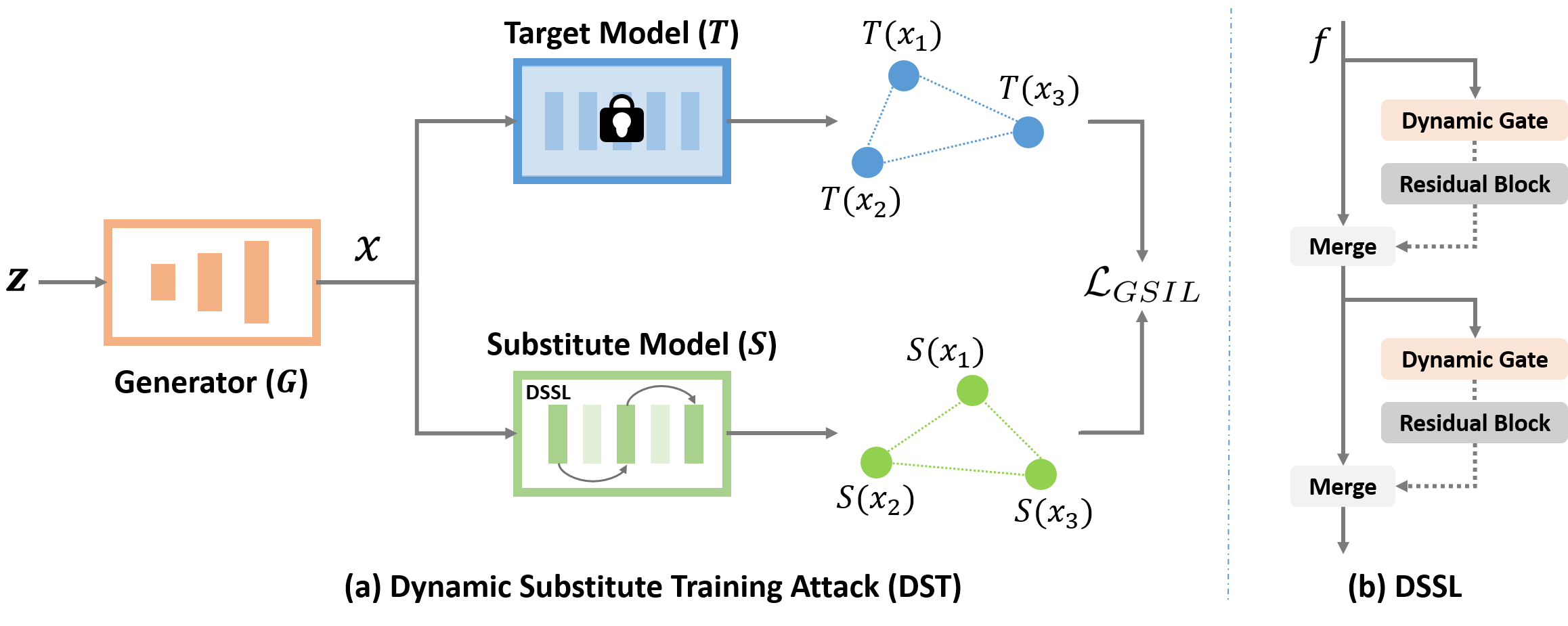}
\par\end{centering}
\caption{(a) Illustration of our \textbf{D}ynamic \textbf{S}ubstitute \textbf{T}raining attack framework (DST). DST utilizes a \textbf{G}raph-based \textbf{S}tructural \textbf{I}nformation \textbf{L}earning constrain $\mathcal{L}_{GSIL}$ to train the generator and substitute model. (b) Schematic diagram of \textbf{D}ynamic \textbf{S}ubstitute \textbf{S}tructure \textbf{L}earning strategy (DSSL). The DSSL of DST is applied to automatically generate optimal substitute model structure according to different targets.
\label{fig:framework}}
\end{figure*}

\noindent \textbf{Data-free Black-box Attack.}
In practice, attackers can hardly obtain the training data for target model, even the number of categories. Thus, some works \cite{mopuri2018generalizable,zhou2020dast,wang2021delving,huan2020data,zhang2021data,kariyappa2021maze} begin to study the data-free black-box attack task, which generate adversarial examples without any knowledge about the training data distribution. Mopuri \emph{et al.} \cite{mopuri2018generalizable} propose an attack to corrupt the extracted features at multiple layers to realize independent of the underlying task. Huan \emph{et al.} \cite{huan2020data} learn adversarial perturbations based on a mapping connection between fine-tuned model and target model. \cite{zhou2020dast,wang2021delving} utilize a generator with noise as input to synthesize data for training a substitute model to learn information from the target one via knowledge distillation. In this paper, we adopt the basic framework as \cite{zhou2020dast,wang2021delving,kariyappa2021maze}. Different from these works, the structure of our substitute model is not fixed, but dynamically generated and optimized according to the different target models and various datasets. Moreover, we train the substitute model and generator not only from a single output of the target model, but also from the detailed and implicit information represented in the graph-based relationship of multiple target outputs.

\section{Methodology}



\subsection{Framework Overview}

Figure~\ref{fig:framework}(a) illustrates the schematic of our proposed unified \textbf{D}ynamic \textbf{S}ubstitute \textbf{T}raining attack framework (DST), which mainly consists of two learnable components, \emph{i.e.}, a generator $\mathcal{G}$ and a substitute model $\mathcal{S}$. More precisely, DST employs a generator $\mathcal{G}$ to synthesize training samples with Gaussian random noise $z \sim \mathcal{N}\left(0,1\right)$, 

\begin{equation}
x = \mathcal{G}\left(z\right) \in \mathbb{R}^{3\times h \times w}
\end{equation}

\noindent where $h$ and $w$ denote the height and width of the generated training samples. Subsequently, we feed the synthesized data into a target model $\mathcal{T}$ and a substitute model $\mathcal{S}$ simultaneously. The teacher-student strategy is re-purposed here to encourage $\mathcal{S}$ to learn as similar decision boundary as $\mathcal{T}$,
\begin{equation}
\mathcal{L}_\mathcal{S} = d\left( \mathcal{T}\left(x\right), \mathcal{S}\left(x\right) \right) 
\label{eq:loss_S}
\end{equation}

\noindent where $d$ denotes a metric function to measure the distance between the outputs from $\mathcal{T}$ and $\mathcal{S}$. After such substitute training process, attacks can be conducted on the well-trained $\mathcal{S}$, and then transferred to $\mathcal{T}$.

As proposed in \cite{zhou2020dast,wang2021delving}, the $\mathcal{S}$ aims to minimize output discrepancy with the $\mathcal{T}$, while the $\mathcal{G}$ tries to maximize the discrepancy to explore various hard samples for substitute training. For model parameters learning via gradient descent, such objective maximizing function for $\mathcal{G}$ learning can be transferred to minimize the loss as the following,
\begin{equation}
\mathcal{L}_\mathcal{G} = - d\left( \mathcal{T}\left(x\right), \mathcal{S}\left(x\right) \right) 
\label{eq:loss_G}
\end{equation}

Overall, the main contributions of this paper concentrate on the learning of substitute model. Remarkably, we for the first time present a dynamic substitute searching strategy to learn an optimal structure for the substitute model, rather than manually select one according to priors~\cite{zhou2020dast,wang2021delving,kariyappa2021maze}. We argue that such a design can best `mimic' the characteristics of the target model, both from the structure of the model and the spatial distribution of parameters, which is more flexible, practical, and significant. Furthermore, in order to stimulate the substitute model to learn more details from the target and improve the quality of generated training data, we propose a graph-based structural information learning strategy to deep explore more structural, relational, and valuable information from multiple target outputs.

\subsection{Dynamic Substitute Structure Learning}
\label{sec:dynamic}
Compared with static network architectures, dynamic network structure usually has the superiority in network capacity to various tasks and targets. In our black-box attack task, according to different target models and various datasets, our \textbf{D}ynamic \textbf{S}ubstitute \textbf{S}tructural \textbf{L}earning strategy (DSSL) can adaptively find more reasonable substitute model architecture to achieve more powerful attack ability. 

Considering that most deep networks have adopted a block-based residual learning design following the remarkable success of ResNet~\cite{he2016deep} (\emph{e.g.}, MobileNetV2~\cite{sandler2018mobilenetv2}, ShuffleNet~\cite{zhang2018shufflenet}, and ResNext~\cite{xie2017aggregated}), 
we, therefore, construct our DSSL based on the residual design to be generally applicable in the common deep networks.

As shown in Fig.~\ref{fig:framework}(b), the selecting probability of each path is generated by a \emph{dynamic gate}. Such dynamic gate aims at predicting a one-hot vector, which denotes whether to execute or skip the branch of residual block. Here, we adapt a set of light-weight operations to realize the dynamic gate function $\mathcal{DG}\left (\cdot  \right )$ with the feature $f$ as the input, 
\begin{equation}
\begin{aligned}
\mathcal{DG}\left ( f\right ) = \mathcal{H}\left ( WP\left ( f \right )+b\right )
\label{eq:defination_gate}
\end{aligned}
\end{equation}
where $P\left (\cdot  \right )$ means the global average pooling layer, and $W$ and $b$ are the fully-connected layer parameters.
To realize discrete binary decisions for the path chosen, we choose hard sigmoid function as $\mathcal{H}\left (\cdot  \right )$, which can be defined as,
\begin{equation}
\begin{aligned}
\mathcal{H}\left ( g\right ) = max \left ( 0, min \left ( kg+\frac{1}{2}, 1\right )\right )
\label{eq:hard_sigmoid}
\end{aligned}
\end{equation}
where we set the threshold as 0.5, which clips the outputs of $\mathcal{H}\left (\cdot  \right )$ to 0 or 1. The $k$ is the crucial parameter used as an approximation of step function that emits binary decisions.

\subsection{Graph-based Structural Information Learning}
\label{sec:graph}

To constrain the consistency of the outputs between the $\mathcal{S}$ and $\mathcal{T}$ and improve the quality of generated training data by $\mathcal{G}$, we argue that it is important to explore the implicit structural relationships among different outputs. The graph-based relational representation can reflect the deep knowledge of $\mathcal{T}$, and learning such structural features can help $\mathcal{S}$ realize a more similar decision boundary as $\mathcal{T}$ to further improve the attack performance. Meanwhile, such structural relationships can reflect the distance among the multiple generated training images, and improve the data quality generated by $\mathcal{G}$ in return.

During the training, we propose a novel \textbf{G}raph-based \textbf{S}tructural \textbf{I}nformation \textbf{L}earning strategy (GSIL) to indict the underlying relationships of the model outputs. By virtue of the graph network conception, in each mini-batch during training, the nodes of graph are the model outputs according to inputs, and the corresponding edges are the relation formulated as an adjacent matrix among these nodes. Specially, the structural information graph is defined as, 
\begin{equation}
\begin{aligned}
Graph = (nodes, edges) = (\left \{ x_{j} \right \}_{j=1}^{B},A) \\
A(j,k) = \left \| x_{j} - x_{k}\right \|_{E}, \quad j,k=1,...,B
\label{eq:defination_graph}
\end{aligned}
\end{equation}
where B means the number of training data in each training iteration, \emph{i.e.}, the value of the mini-batch, and the edge is defined as the Euclidean distance $E$ between the two node representations / model outputs. 

Once we obtain the structural information graph, the corresponding constrain $\mathcal{L}_{GSIL}$ can be produced to further restrict the discrepancy between the $Graph^{S}$ of $\mathcal{S}$ and the $Graph^{T}$ of $\mathcal{T}$. The difference of such graphs contains the node discrepancy and edge discrepancy formulated as,
\begin{small}
\begin{equation}
\begin{aligned}
\mathcal{L}_{GSIL} & =Disc(Graph^{T},Graph^{S})\\
 & =\alpha_{1}\sum_{j=0}^{B}\cdot KL\left(x_{j}^{T},x_{j}^{S}\right)+\alpha_{2}\cdot MSE\left(A^{T},A^{S}\right)
 \label{eq:loss_GSIL}
\end{aligned}
\end{equation}
\end{small}
where $x_{j}^{T},x_{j}^{S}$ and $A^{T},A^{S}$ refer to the node and edge set for $\mathcal{S}$ and $\mathcal{T}$. The Kullback-Leibler Divergence function $KL$ is used to measure the discrepancy between two nodes, and the MSE loss is applied to restrict the differences between the two edges. And $\alpha_{1}$ and $\alpha_{2}$ are the hyper-parameters to balance the discrepancy of the nodes and edges. In DST model learning, we apply the $\mathcal{L}_{GSIL}$ to measure the discrepancy between the $\mathcal{S}$ and $\mathcal{T}$ used in Eq.~\ref{eq:loss_S} and Eq.~\ref{eq:loss_G}.

\begin{table*}
\begin{spacing}{0.9}
\begin{centering}
\setlength{\tabcolsep}{1.6mm}{
\begin{tabular} {c}
\hspace{-0.1in}
\begin{tabular}{c|c|ccc|ccc|cc|cc}
\hline 
 & {\small{}Dataset } & \multicolumn{3}{c|}{{\small{}MNIST }} & \multicolumn{3}{c|}{{\small{}CIFAR-10 }} & \multicolumn{2}{c|}{{\small{}CIFAR-100 }} & \multicolumn{1}{c}{{\small{}Tiny ImageNet}}\tabularnewline
\hline 
 & {\small{}Target Model} & {\small{}AlexNet} & {\small{}VGG-16} & {\small{}ResNet-18} & {\small{}AlexNet} & {\small{}VGG-16} & {\small{}ResNet-18} & {\small{}VGG-19} & {\small{}ResNet-50} & {\small{}ResNet-50}\tabularnewline
\hline 
\hline 
\multirow{6}{*}{\begin{turn}{90}{\small{}Non-Target} \end{turn}} 
& {\small{}GD-UAP \cite{mopuri2018generalizable}} & {\small{}33.28} & {\small{}29.54} & {\small{}30.81} & {\small{}18.39} & {\small{}16.43} & {\small{}20.65} & {\small{}12.57} & {\small{}14.90} & {\small{}8.93} \tabularnewline
& {\small{}Cosine-UAP \cite{zhang2021data}} & {\small{}38.92} & {\small{}35.11} & {\small{}28.48} & {\small{}38.20} & {\small{}23.44} & {\small{}35.73} & {\small{}15.62} & {\small{}17.83} & {\small{}11.96} \tabularnewline
\cline{2-11}
& {\small{}DaST \cite{zhou2020dast}} & {\small{}63.34} & {\small{}60.38} & {\small{}56.21} & {\small{}43.64} & {\small{}56.25} & {\small{}49.36} & {\small{}32.94} & {\small{}27.32} & {\small{}26.85} \tabularnewline 
& {\small{}MAZE \cite{kariyappa2021maze}} & {\small{}65.82} & {\small{}67.68} & {\small{}59.32} & {\small{}44.72} & {\small{}50.13} & {\small{}52.99} & {\small{}29.41} & {\small{}24.83} & {\small{}25.40} \tabularnewline
& {\small{}DDG+AST \cite{wang2021delving}} & {\small{}68.29} & {\small{}65.03} & {\small{}61.47} & {\small{}44.87} & {\small{}53.91} & {\small{}50.30} & {\small{}31.88} & {\small{}26.56} & {\small{}30.81} \tabularnewline
\cline{2-11}
& \small{}DST (Ours) & \textbf{\small{}70.48} & \textbf{\small{}72.49} & \textbf{\small{}63.72} & \textbf{\small{}47.20} & \textbf{\small{}58.21} & \textbf{\small{}54.93} & \textbf{\small{}34.03} & \textbf{\small{}31.39} & \textbf{\small{}32.28} \tabularnewline
\hline 
\hline 
\multirow{6}{*}{\begin{turn}{90}{\small{}Target} \end{turn}} 
& {\small{}GD-UAP \cite{mopuri2018generalizable}} & {\small{}28.10} & {\small{}39.51} & {\small{}29.48} & {\small{}14.22} & {\small{}18.43} & {\small{}16.49} & {\small{}10.55} & {\small{}6.31} & {\small{}7.50} \tabularnewline
& {\small{}Cosine-UAP \cite{zhang2021data}} & {\small{}36.91} & {\small{}48.23} & {\small{}37.88} & {\small{}20.46} & {\small{}17.97} & {\small{}24.31} & {\small{}12.40} & {\small{}12.11} & {\small{}10.53} \tabularnewline
\cline{2-11}
& {\small{}DaST \cite{zhou2020dast}} & {\small{}58.28} & {\small{}67.33} & {\small{}54.29} & {\small{}29.48} & {\small{}40.29} & {\small{}46.10} & {\small{}15.82} & {\small{}22.48} & {\small{}20.37} \tabularnewline 
& {\small{}MAZE \cite{kariyappa2021maze}} & {\small{}60.48} & {\small{}67.39} & {\small{}60.43} & {\small{}33.28} & {\small{}29.83} & {\small{}41.26} & {\small{}18.22} & {\small{}20.21} & {\small{}19.25} \tabularnewline
& {\small{}DDG+AST \cite{wang2021delving}} & {\small{}62.20} & {\small{}66.45} & {\small{}61.94} & {\small{}35.91} & {\small{}34.87} & {\small{}45.25} & {\small{}17.04} & {\small{}19.57} & {\small{}17.48} \tabularnewline
\cline{2-11}
& \small{}DST (Ours) & \textbf{\small{}64.82} & \textbf{\small{}68.49} & \textbf{\small{}65.77} & \textbf{\small{}38.29} & \textbf{\small{}44.71} & \textbf{\small{}47.90} & \textbf{\small{}20.59} & \textbf{\small{}23.01} & \textbf{\small{}22.94} \tabularnewline
\hline 
\end{tabular}
\end{tabular}}
\caption{Comparing ASRs results using probability as the target model output among our method and competitors over four datasets. For a fair comparison, we use PGD as the attack method and ResNet-34 as the default substitute model for all substitute training.
\label{Tab: Comparsion-P}}
\par\end{centering}
\end{spacing}
\end{table*}

\begin{table*}
\begin{spacing}{0.9}
\begin{centering}
\setlength{\tabcolsep}{2.2mm}{
\begin{tabular} {c}
\hspace{-0.1in}
\begin{tabular}{c|c|cccc|cccc|cccc}
\hline 
 & {\small{}Dataset } & \multicolumn{4}{c|}{{\small{}MNIST }} & \multicolumn{4}{c|}{{\small{}CIFAR-10 }} & \multicolumn{4}{c}{{\small{}CIFAR-100 }} \tabularnewline
\hline 
 & {\small{}Attack Method} & {\small{}FGSM} & {\small{}BIM} & {\small{}PGD} & {\small{}C\&W} & {\small{}FGSM} & {\small{}BIM} & {\small{}PGD} & {\small{}C\&W} & {\small{}FGSM} & {\small{}BIM} & {\small{}PGD} & {\small{}C\&W}  \tabularnewline
\hline 
\hline 
\multirow{4}{*}{\begin{turn}{90}{\small{}Non-Target} \end{turn}} 
& {\small{}DaST \cite{zhou2020dast}} & {\small{}59.32} & {\small{}81.52} & {\small{}63.34} & {\small{}59.37} & {\small{}42.41} & {\small{}53.92} & {\small{}56.25} & {\small{}57.28} & {\small{}27.48} & {\small{}34.56} & {\small{}27.32} & {\small{}25.37} \tabularnewline
& {\small{}MAZE \cite{kariyappa2021maze}} & {\small{}56.25} & {\small{}74.83} & {\small{}65.82} & {\small{}70.44} & {\small{}46.57} & {\small{}62.91} & {\small{}50.13} & {\small{}48.12} & {\small{}27.86} & {\small{}30.51} & {\small{}24.83} & {\small{}27.66} \tabularnewline
& {\small{}DDG+AST \cite{wang2021delving}} & {\small{}62.48} & {\small{}76.70} & {\small{}68.29} & {\small{}69.33} & {\small{}44.12} & {\small{}59.03} & {\small{}53.91} & {\small{}56.28} & {\small{}30.79} & {\small{}33.62} & {\small{}26.56} & {\small{}25.83} \tabularnewline
\cline{2-14}
& {\small{}DST (Ours)} & {\small{}\textbf{63.92}} & {\small{}\textbf{82.45}} & {\small{}\textbf{70.48}} & {\small{}\textbf{73.22}} & {\small{}\textbf{49.21}} & {\small{}\textbf{64.90}} & {\small{}\textbf{58.21}} & {\small{}\textbf{59.17}} & {\small{}\textbf{33.80}} & {\small{}\textbf{37.74}} & {\small{}\textbf{31.39}} & {\small{}\textbf{29.33}} \tabularnewline
\hline 
\hline 
\multirow{4}{*}{\begin{turn}{90}{\small{}Target} \end{turn}} 
& {\small{}DaST \cite{zhou2020dast}} & {\small{}59.28} & {\small{}70.47} & {\small{}58.28} & {\small{}59.33} & {\small{}30.24} & {\small{}45.10} & {\small{}40.29} & {\small{}42.13} & {\small{}16.49} & {\small{}\textbf{27.86}} & {\small{}22.48} & {\small{}24.56} \tabularnewline
& {\small{}MAZE \cite{kariyappa2021maze}} & {\small{}61.42} & {\small{}67.29} & {\small{}60.48} & {\small{}57.32} & {\small{}26.35} & {\small{}40.49} & {\small{}29.83} & {\small{}38.10} & {\small{}20.42} & {\small{}24.81} & {\small{}20.21} & {\small{}23.85} \tabularnewline
& {\small{}DDG+AST \cite{wang2021delving}} & {\small{}57.84} & {\small{}71.90} & {\small{}62.20} & {\small{}52.11} & {\small{}37.58} & {\small{}42.06} & {\small{}34.87} & {\small{}45.39} & {\small{}17.82} & {\small{}26.33} & {\small{}19.57} & {\small{}28.95} \tabularnewline
\cline{2-14}
& {\small{}DST (Ours)} & {\small{}\textbf{64.23}} & {\small{}\textbf{73.85}} & {\small{}\textbf{64.82}} & {\small{}\textbf{60.57}} & {\small{}\textbf{39.53}} & {\small{}\textbf{47.20}} & {\small{}\textbf{44.71}} & {\small{}\textbf{48.06}} & {\small{}\textbf{20.48}} & {\small{}27.31} & {\small{}\textbf{23.01}} & {\small{}\textbf{29.57}} \tabularnewline
\hline 
\end{tabular}
\end{tabular}}
\caption{Comparing ASRs results using probability as the target model output among our method and competitors with various white-box adversarial example generation methods. For a fair comparison, we utilize ResNet-34 as the substitute model for all substitute training. The target models are the AlexNet for MNIST, VGG-16 for CIFAR-10, and ResNet-50 for CIFAR-100.
\label{Tab: Comparsion-attacks}}
\par\end{centering}
\end{spacing}
\end{table*}

\section{Experiment}
\subsection{Experiment Setup}
\noindent \textbf{Datasets and model structure.}
1) MNIST~\cite{lecun1998gradient}: 
The target model is pre-trained on AlexNet~\cite{krizhevsky2017imagenet}, VGG-16~\cite{simonyan2014very}, and ResNet-18~\cite{he2016deep}.
2) CIFAR-10~\cite{krizhevsky2009learning}: 
The target model is pre-trained on AlexNet, VGG-16, and ResNet-18.
3) CIFAR-100~\cite{krizhevsky2009learning}: 
The target model is pre-trained on VGG-19 and ResNet-50.
4) Tiny Imagenet~\cite{russakovsky2015imagenet}: 
The target model is pre-trained on ResNet-50. 
For these four datasets, the default substitute model basic structure is ResNet-34. 

\noindent \textbf{Competitors.} To verify the efficacy of our DST, we compare our attacking results with the existing state-of-the-art data-free black-box attacks, \textit{i.e.}, GD-UAP~\cite{mopuri2018generalizable}, Cosine-UAP~\cite{zhang2021data}, DaST~\cite{zhou2020dast}, MAZE~\cite{kariyappa2021maze}, and DDG+AST~\cite{wang2021delving}.

\noindent \textbf{Implementation details.}
We use Pytorch for Implementation. We utilize Adam to train our substitute model and generator from scratch, and all weights are randomly initialized using a truncated normal distribution with std of 0.02. The initial learning rates of the generator and substitute model are set as 0.0001 and 0.001, respectively, they are gradually decreased to zero from the 80th epoch, and stopped at the 150th epoch. We set the mini-batch size as 500, the hyper-parameters $ \alpha_1$ and $\alpha_2$ are equally as 1. The $k$ in Eq.~\ref{eq:hard_sigmoid} is set as 1 in the following experiments. Our model is trained by one NVIDIA GeForce GTX 1080Ti GPU. We apply PGD~\cite{madry2017pgd} as the default white-box attack method to generate adversarial images over the well-trained substitute model during the evaluation. We also utilize several classic attack methods for extensive experiments, such as FGSM~\cite{goodfellow2014explaining}, BIM~\cite{kurakin2016adversarial} and C\&W~\cite{carlini2017CW}. During the model optimizing stage, there are no real images used for model learning, only random noise is utilized as input. For evaluation, the adversarial samples to conduct attack crafted only on test set over four datasets.


\noindent \textbf{Evaluation metrics.}
Following the two scenarios proposed in DaST \cite{zhou2020dast}, \textit{i.e.}, only getting the output label from the target model and accessing the output probability well, we name these two scenarios as Probability-based and Label-based. In the experiments, we report the attack success rates (ASRs) of the adversarial examples generated by the substitute model to attack the target model. As in DaST \cite{zhou2020dast}, for non-target attacking, we only attack the images classified correctly by the target model. For target attacking, we only generate adversarial examples on the images which are not classified to the specific wrong labels. 
We conduct ten times over each testing, and report the average results.

\begin{table*}
\begin{centering}
\setlength{\tabcolsep}{0.9mm}{
\begin{tabular} {c}
\hspace{-0.1in}
\begin{tabular}{c|c|ccccc|ccccc|ccccc}
\hline 
\multirow{2}{*}{} & \multirow{2}{*}{Method} & \multicolumn{5}{c|}{MNIST}  & \multicolumn{5}{c|}{CIFAR-10} & \multicolumn{5}{c}{CIFAR-100} \tabularnewline
\cline{3-17}
 &  & V-16 & V-19 & R-18 & R-34 & R-50 & V-16 & V-19 & R-18 & R-34 & R-50 & V-16 & V-19 & R-18 & R-34 & R-50 \tabularnewline
\hline 
\multirow{4}{*}{\begin{turn}{90} Non-Target \end{turn}}
& DaST \cite{zhou2020dast} & \textcolor{blue}{62.49} & 59.21 & 40.36 & 60.38 & 49.85 & 23.48 & 45.83 & 34.87 & \textcolor{blue}{49.36} & 43.20 & 17.38 & 12.42 & 20.48 & \textcolor{blue}{27.32} & 22.56 \tabularnewline
& MAZE \cite{kariyappa2021maze} & 57.36 & \textcolor{blue}{69.31} & 64.20 & 67.68 & 55.47 & 36.88 & 29.42 & 38.47 & \textcolor{blue}{52.99} & 50.35 & 13.27 & 23.49 & 19.21 & 24.83 & \textcolor{blue}{25.55} \tabularnewline
& DDG+AST \cite{wang2021delving} & 68.28 & 57.20 & 63.12 & 65.03 & \textcolor{blue}{68.37} & 36.02 & \textcolor{blue}{52.30} & 47.21 & 50.30 & 49.46 & 12.48 & 17.20 & 15.93 & \textcolor{blue}{26.56} & 22.46 \tabularnewline
\cline{2-17}
& DST (Ours) & \multicolumn{5}{c|}{\textbf{72.49}} & \multicolumn{5}{c|}{\textbf{54.93}} & \multicolumn{5}{c}{\textbf{31.39}} \tabularnewline 
\hline 
\hline 
\multirow{4}{*}{\begin{turn}{90} Target \end{turn}}
& DaST \cite{zhou2020dast} & 53.92 & 61.48 & 42.01 & \textcolor{blue}{67.33} & 48.47 & 19.49 & 39.61 & 37.02 & 46.10 & \textcolor{blue}{47.00} & 9.32 & \textcolor{blue}{22.51} & 14.29 & 22.48 & 17.56 \tabularnewline
& MAZE \cite{kariyappa2021maze} & 54.21 & \textcolor{blue}{67.50} & 57.86 & 67.39 & 64.23 & \textcolor{blue}{42.93} & 35.32 & 37.48 & 41.26 & 36.06 & 12.41 & \textcolor{blue}{21.48} & 18.60 & 20.21 & 20.35 \tabularnewline
& DDG+AST \cite{wang2021delving} & 48.99 & 63.81 & \textcolor{blue}{66.91} & 66.45 & 55.30 & 43.05 & 41.29 & 40.82 & \textcolor{blue}{45.25} & 38.90 & 13.40 & 18.66 & \textcolor{blue}{21.43} & 19.57 & 19.95 \tabularnewline
\cline{2-17}
& DST (Ours) & \multicolumn{5}{c|}{\textbf{68.49}} & \multicolumn{5}{c|}{\textbf{47.90}} & \multicolumn{5}{c}{\textbf{23.01}} \tabularnewline 
\hline 
\end{tabular}
\end{tabular}}
\caption{ASRs comparisons among our DST and competitors using different substitute model structures under Probability-based scenario. The target model are VGG-16 for MNIST, ResNet-18 for CIFAR-10, and ResNet-50 for CIFAR-100. We use PGD as the attack method for all attacks. The competitors can use VGG-16 (V-16), VGG-19 (V-19), ResNet-18 (R-18), ResNet-34 (R-34), and ResNet-50 (R-50) as their substitute models, and the ASRs with blue represent the selected best substitute models according to different attack targets for each competitor. Our DST attack can only use ResNet-34 as the basic substitute model structure under all conditions.
\label{Tab: Abaltion-structures}}
\par\end{centering}
\end{table*}

\begin{table}
\begin{centering}
\setlength{\tabcolsep}{1.6mm}{
\begin{tabular} {c}
\hspace{-0.1in}
\begin{tabular}{c|c|cc}
\hline 
 & Method & Probability-based & Lable-based\tabularnewline
\hline 
\multirow{6}{*}{\begin{turn}{90}{\small{}Non-Target}\end{turn}} 
& GD-UAP \cite{mopuri2018generalizable} & 73.12 & 68.01 \tabularnewline
& Cosine-UAP \cite{zhang2021data} & 77.48 & 82.56 \tabularnewline
\cline{2-4}
& DaST \cite{zhou2020dast} & 92.48 & 94.21 \tabularnewline
& MAZE \cite{kariyappa2021maze} & 93.95 & 93.18 \tabularnewline
& DDG+AST \cite{wang2021delving} & 93.29 & 95.83 \tabularnewline
\cline{2-4}
& DST (Ours) & \textbf{95.02} & \textbf{96.44} \tabularnewline
\hline 
\multirow{6}{*}{\begin{turn}{90}{\small{}Target}\end{turn}} 
& GD-UAP \cite{mopuri2018generalizable} & 42.19 & 31.47 \tabularnewline
& Cosine-UAP \cite{zhang2021data} & 46.10 & 52.36 \tabularnewline
\cline{2-4}
& DaST \cite{zhou2020dast} & 52.18 & 68.85 \tabularnewline
& MAZE \cite{kariyappa2021maze} & 50.23 & 59.83 \tabularnewline
& DDG+AST \cite{wang2021delving} & 53.46 & 71.48 \tabularnewline
\cline{2-4}
& DST (Ours) & \textbf{56.39} & \textbf{75.90} \tabularnewline
\hline 
\end{tabular}
\end{tabular}}
\caption{Comparing ASRs results among our proposed DST attack and competitors for attacking the online Microsoft Azure example model under both Probability- and Label-based scenarios. For a fair comparison, we use PGD as the attack method and ResNet-34 as the default substitute model for all substitute training.
\label{Tab: Comparsion-Azure}}
\par\end{centering}
\end{table}

\subsection{Black-box Attack Results}

\noindent \textbf{Comparisons with the state-of-the-art attacks.}
As shown in Tab.~\ref{Tab: Comparsion-P}, our DST attack beats all competitors with significant margins. We conduct extensive comparisons with these competitors from several aspects, \emph{i.e.}, the diverse tasks (datasets), various target models, and different attack goals (target/non-target). The results show that our DST method can achieve the best attack ability for black-box attack without using any real data as input.

\noindent \textbf{Comparisons with competitors using various white-box adversarial sample generation methods.}
Under the probability-based scenario, we compare the ASRs with competitors when using different white-box attack methods via well-trained substitute models. As illustrated in Tab.~\ref{Tab: Comparsion-attacks}, we apply four classic attacks to generate adversarial samples over substitute models for attacking the target models on three datasets. Compared to other data-free black-box attack algorithms, our DST can dynamically generate suitable substitute model structure and learn structural information from the target output, thus, DST achieves the best ASRs over most experiments.

\noindent \textbf{Comparisons with competitors which use different deep networks as their substitute model.}
As shown in Tab.~\ref{Tab: Abaltion-structures}, we compare our DST attack performance with competitors which can choose a bunch of networks as their substitute model. Even our DST can only use ResNet-34 as the basic substitute model, thanks to the dynamic substitute structure learning strategy, we can still achieve better attack performance when the competitors are able to choose the best substitute model from a set of different networks. We can also notice that the attack performance of the competitors is highly dependent on the chosen substitute model structure. Such results demonstrate that our DST can automatically generate optimal substitute model according to the target, which is crucial for the practical applications.

\noindent \textbf{Comparisons with state-of-the-art competitors against online Microsoft Azure example model.}
The attack performance targeting real-world black-box model is vital to evaluate the adversarial example generation method. Thus, we compare our DST with competitors over the online Microsoft Azure model. As in Tab.~\ref{Tab: Comparsion-Azure}, our DST outperforms all competitors, which indicates the practical black-box attack capacity of the proposed DST in real-world applications.

\begin{table}
\begin{centering}
\setlength{\tabcolsep}{1.4mm}{
\begin{tabular} {c}
\hspace{-0.1in}
\begin{tabular}{c|c|cc|cc}
\hline 
\multirow{2}{*}{} & \multirow{2}{*}{Components} & \multicolumn{2}{c|}{Probability-based} & \multicolumn{2}{c}{Label-based}\tabularnewline
\cline{3-6}
 &  & MNIST & C-100 & MNIST & C-100\tabularnewline
\hline 
\multirow{4}{*}{\begin{turn}{90} Non-Target \end{turn}}
& Baseline-I & 30.82 & 12.40 & 15.33 & 7.54 \tabularnewline
& Baseline-II & 48.99 & 21.36 & 23.47 & 14.22 \tabularnewline
& + GSIL & 59.32 & 24.81 & 29.65 & 20.23 \tabularnewline
& + DSSL & \textbf{70.48} & \textbf{31.39} & \textbf{36.22} & \textbf{25.83} \tabularnewline 
\hline 
\hline 
\multirow{4}{*}{\begin{turn}{90} Target \end{turn}}
& Baseline-I & 27.49 & 10.48 & 16.72 & 5.83 \tabularnewline
& Baseline-II & 38.21 & 18.49 & 23.95 & 8.49 \tabularnewline
& + GSIL & 52.40 & 21.48 & 26.58 & 13.77 \tabularnewline
& + DSSL & \textbf{64.82} & \textbf{23.01} & \textbf{31.94} & \textbf{19.30} \tabularnewline 
\hline 
\end{tabular}
\end{tabular}}
\caption{ASRs results of variants of the proposed DST attack method. 
The target model is based on AlexNet for MNIST, and ResNet-50 for CIFAR-100. `C-100' refers to the CIFAR-100 dataset. We use PGD as the attack method and ResNet-34 as the substitute model for all experiments.
\label{Tab: Abaltion}}
\par\end{centering}
\end{table}

\subsection{Ablation Study}
To further explore the efficacy of components in our DST attack, we conduct extensive ablation studies over the following variants: (1) `Baseline-I': using random noise as input to generate training data and applying MSE loss to constrain the outputs' similarity between the $\mathcal{S}$ and $\mathcal{T}$; (2) `Baseline-II': using random noise as input to generate training data and applying Kullback-Leibler Divergence to constrain the outputs' similarity between the $\mathcal{S}$ and $\mathcal{T}$; (3) `+ GSIL': using the graph-based structural information learning constrain to conduct substitute training; (4) `+ DSSL': based on the `+ GSIL', a dynamic substitute structure learning strategy is applied to adaptively generate suitable substitute model structure, and it is the whole DST attack model.

\begin{table*}[h]
\begin{spacing}{0.9}
\begin{centering}
\setlength{\tabcolsep}{1mm}{
\begin{tabular} {c}
\hspace{-0.1in}
\begin{tabular}{c|c|ccc|ccc|cc}
\hline 
 & {\small{}Dataset } & \multicolumn{3}{c|}{{\small{}MNIST }} & \multicolumn{3}{c|}{{\small{}CIFAR-10 }} & \multicolumn{2}{c}{{\small{}CIFAR-100 }} \tabularnewline
\hline 
 & {\small{}Target Model} & {\small{}AlexNet} & {\small{}VGG-16} & {\small{}ResNet-18} & {\small{}AlexNet} & {\small{}VGG-16} & {\small{}ResNet-18} & {\small{}VGG-19} & {\small{}ResNet-50} \tabularnewline
\hline 
\hline 
\multirow{8}{*}{\begin{turn}{90}{\small{}Non-Target} \end{turn}} 
& {\small{}ResNet-18} & {\small{}59.41} & {\small{}49.82} & {\small{}48.24} & {\small{}35.67} & {\small{}54.28} & {\small{}43.09} & {\small{}12.49} & {\small{}14.23} \tabularnewline
& {\small{} \cellcolor[HTML]{defcde} + DSSL (Ours)} & {\small{} \cellcolor[HTML]{defcde} 68.81(\textcolor{blue}{55.6})} & {\small{} \cellcolor[HTML]{defcde} 73.59(\textcolor{blue}{50.0})} & {\small{} \cellcolor[HTML]{defcde} 60.40(\textcolor{blue}{44.4})} & {\small{} \cellcolor[HTML]{defcde} 50.91(\textcolor{blue}{38.9})} & {\small{} \cellcolor[HTML]{defcde} 57.24(\textcolor{blue}{33.3})} & {\small{} \cellcolor[HTML]{defcde} 55.96(\textcolor{blue}{22.2})} & {\small{} \cellcolor[HTML]{defcde} 30.43(\textcolor{blue}{22.2})} & {\small{} \cellcolor[HTML]{defcde} 27.98(\textcolor{blue}{11.1})} \tabularnewline
\cline{2-10}
& {\small{}ResNet-34} & {\small{}59.32} & {\small{}64.40} & {\small{}56.28} & {\small{}39.21} & {\small{}50.90} & {\small{}46.29} & {\small{}23.55} & {\small{}24.81} \tabularnewline
& {\small{} \cellcolor[HTML]{defcde} + DSSL (Ours)} & {\small{} \cellcolor[HTML]{defcde} 70.48(\textcolor{blue}{70.6})} & {\small{} \cellcolor[HTML]{defcde} 72.49(\textcolor{blue}{73.5})} & {\small{} \cellcolor[HTML]{defcde} 63.72(\textcolor{blue}{55.9})} & {\small{} \cellcolor[HTML]{defcde} 47.20(\textcolor{blue}{61.8})} & {\small{} \cellcolor[HTML]{defcde} 58.21(\textcolor{blue}{55.9})} & {\small{} \cellcolor[HTML]{defcde} 54.93(\textcolor{blue}{67.7})} & {\small{} \cellcolor[HTML]{defcde} 34.03(\textcolor{blue}{55.9})} & {\small{} \cellcolor[HTML]{defcde} 31.39(\textcolor{blue}{38.2})} \tabularnewline
\cline{2-10}
& {\small{}ResNet-50} & {\small{}47.80} & {\small{}63.36} & {\small{}50.19} & {\small{}38.42} & {\small{}43.28} & {\small{}50.12} & {\small{}29.65} & {\small{}17.33} \tabularnewline
& {\small{} \cellcolor[HTML]{defcde} + DSSL (Ours)} & {\small{} \cellcolor[HTML]{defcde} 69.55(\textcolor{blue}{86.0})} & {\small{} \cellcolor[HTML]{defcde} 73.42(\textcolor{blue}{80.0})} & {\small{} \cellcolor[HTML]{defcde} 61.82(\textcolor{blue}{82.0})} & {\small{} \cellcolor[HTML]{defcde} 50.37(\textcolor{blue}{78.0})} & {\small{} \cellcolor[HTML]{defcde} 55.61(\textcolor{blue}{60.0})} & {\small{} \cellcolor[HTML]{defcde} 53.57(\textcolor{blue}{70.0})} & {\small{} \cellcolor[HTML]{defcde} 33.50(\textcolor{blue}{42.0})} & {\small{} \cellcolor[HTML]{defcde} 32.40(\textcolor{blue}{54.0})} \tabularnewline
\cline{2-10} 
& {\small{}ResNet-101} & {\small{}49.08} & {\small{}58.21} & {\small{}38.17} & {\small{}32.54} & {\small{}54.83} & {\small{}49.22} & {\small{}19.34} & {\small{}23.46} \tabularnewline
& {\small{} \cellcolor[HTML]{defcde} + DSSL (Ours)} & {\small{} \cellcolor[HTML]{defcde} 70.33(\textcolor{blue}{92.1})} & {\small{} \cellcolor[HTML]{defcde} 74.02(\textcolor{blue}{87.1})} & {\small{} \cellcolor[HTML]{defcde} 59.48(\textcolor{blue}{93.1})} & {\small{} \cellcolor[HTML]{defcde} 49.53(\textcolor{blue}{88.1})} & {\small{} \cellcolor[HTML]{defcde} 57.31(\textcolor{blue}{88.1})} & {\small{} \cellcolor[HTML]{defcde} 56.24(\textcolor{blue}{79.2})} & {\small{} \cellcolor[HTML]{defcde} 35.09(\textcolor{blue}{67.3})} & {\small{} \cellcolor[HTML]{defcde} 30.47(\textcolor{blue}{77.2})} \tabularnewline
\cline{2-10} 
\hline 
\hline 
\multirow{8}{*}{\begin{turn}{90}{\small{}Non-Target} \end{turn}} 
& {\small{}ResNet-18} & {\small{}60.35} & {\small{}58.28} & {\small{}59.12} & {\small{}28.44} & {\small{}43.75} & {\small{}49.20} & {\small{}15.91} & {\small{}11.27} \tabularnewline
& {\small{} \cellcolor[HTML]{defcde} + DSSL (Ours)} & {\small{} \cellcolor[HTML]{defcde} 62.94(\textcolor{blue}{55.6})} & {\small{} \cellcolor[HTML]{defcde} 67.21(\textcolor{blue}{50.0})} & {\small{} \cellcolor[HTML]{defcde} 63.38(\textcolor{blue}{44.4})} & {\small{} \cellcolor[HTML]{defcde} 39.41(\textcolor{blue}{38.9})} & {\small{} \cellcolor[HTML]{defcde} 46.82(\textcolor{blue}{33.3})} & {\small{} \cellcolor[HTML]{defcde} 48.03(\textcolor{blue}{22.2})} & {\small{} \cellcolor[HTML]{defcde} 16.23(\textcolor{blue}{22.2})} & {\small{} \cellcolor[HTML]{defcde} 21.10(\textcolor{blue}{11.1})} \tabularnewline
\cline{2-10} 
& {\small{}ResNet-34} & {\small{}52.40} & {\small{}60.22} & {\small{}55.94} & {\small{}32.51} & {\small{}35.94} & {\small{}39.42} & {\small{}14.09} & {\small{}21.48} \tabularnewline
& {\small{} \cellcolor[HTML]{defcde} + DSSL (Ours)} & {\small{} \cellcolor[HTML]{defcde} 64.82(\textcolor{blue}{70.6})} & {\small{} \cellcolor[HTML]{defcde} 68.49(\textcolor{blue}{73.5})} & {\small{} \cellcolor[HTML]{defcde} 65.77(\textcolor{blue}{55.9})} & {\small{} \cellcolor[HTML]{defcde} 38.29(\textcolor{blue}{61.8})} & {\small{} \cellcolor[HTML]{defcde} 44.71(\textcolor{blue}{55.9})} & {\small{} \cellcolor[HTML]{defcde} 47.90(\textcolor{blue}{67.7})} & {\small{} \cellcolor[HTML]{defcde} 20.59(\textcolor{blue}{55.9})} & {\small{} \cellcolor[HTML]{defcde} 23.01(\textcolor{blue}{38.2})} \tabularnewline
\cline{2-10} 
& {\small{}ResNet-50} & {\small{}54.28} & {\small{}63.47} & {\small{}60.02} & {\small{}33.00} & {\small{}41.25} & {\small{}35.91} & {\small{}17.32} & {\small{}23.99} \tabularnewline
& {\small{} \cellcolor[HTML]{defcde} + DSSL (Ours)} & {\small{} \cellcolor[HTML]{defcde} 66.91(\textcolor{blue}{86.0})} & {\small{} \cellcolor[HTML]{defcde} 70.36(\textcolor{blue}{80.0})} & {\small{} \cellcolor[HTML]{defcde} 64.24(\textcolor{blue}{82.0})} & {\small{} \cellcolor[HTML]{defcde} 37.58(\textcolor{blue}{78.0})} & {\small{} \cellcolor[HTML]{defcde} 43.92(\textcolor{blue}{60.0})} & {\small{} \cellcolor[HTML]{defcde} 49.50(\textcolor{blue}{70.0})} & {\small{} \cellcolor[HTML]{defcde} 21.42(\textcolor{blue}{42.0})} & {\small{} \cellcolor[HTML]{defcde} 25.84(\textcolor{blue}{54.0})} \tabularnewline
\cline{2-10} 
& {\small{}ResNet-101} & {\small{}43.35} & {\small{}56.17} & {\small{}63.29} & {\small{}30.88} & {\small{}37.25} & {\small{}42.63} & {\small{}21.38} & {\small{}14.56} \tabularnewline
& {\small{} \cellcolor[HTML]{defcde} + DSSL (Ours)} & {\small{} \cellcolor[HTML]{defcde} 64.27(\textcolor{blue}{92.1})} & {\small{} \cellcolor[HTML]{defcde} 66.05(\textcolor{blue}{87.1})} & {\small{} \cellcolor[HTML]{defcde} 64.88(\textcolor{blue}{93.1})} & {\small{} \cellcolor[HTML]{defcde} 40.10(\textcolor{blue}{88.1})} & {\small{} \cellcolor[HTML]{defcde} 41.29(\textcolor{blue}{88.1})} & {\small{} \cellcolor[HTML]{defcde} 48.57(\textcolor{blue}{79.2})} & {\small{} \cellcolor[HTML]{defcde} 22.50(\textcolor{blue}{67.3})} & {\small{} \cellcolor[HTML]{defcde} 23.36(\textcolor{blue}{77.2})} \tabularnewline
\cline{2-10} 
\hline 
\end{tabular}
\end{tabular}}
\caption{Comparing ASRs results using probability as the target output among our method and variants with different substitute networks. We use PGD as the default attack for all experiments. The `ResNet-18', `ResNet-34', `ResNet-50', and `ResNet-101' represent the basic substitute model with graph-based structural information learning module, and `+DSSL (ours)' means the above raw model with dynamic substitute structure learning strategy. The numbers in `()' denote \textcolor{blue}{skip rate} (the percentage (\%) of skip blocks in the substitute model).
\label{Tab: Comparsion-DSSL}}
\par\end{centering}
\end{spacing}
\end{table*}

\noindent \textbf{The efficacy of the components for attack ability in DST attack.}
As in Tab.~\ref{Tab: Abaltion}, comparing the results among the variants, we can notice the following observations,
(1) `Baseline-I' can only realize basic attack ability against the target model.
(2) Comparing the `Baseline-II' with `Baseline-I', it is obvious that changing the MSE loss to Kullback-Leibler Divergence constrain can better restrict the output similarity between the $\mathcal{S}$ and $\mathcal{T}$.
(3) Comparing the results between the `Baseline-II' and `+ GSIL', the effectiveness of the structural information learning strategy is proved. Specially, the distance of graph edges represent the outputs' structural relationship, thus, learning such information can greatly encourage $\mathcal{S}$ to learn more detailed knowledge from the $\mathcal{T}$.
(4) With the `+DSSL' module, the ASRs have been significantly improved. Such results illustrate that the static substitute structure does not benefit to various attack targets, and self-optimized substitute structure is more reasonable for unknown diverse attack targets.

\noindent \textbf{The efficacy of dynamic substitute structure learning strategy for different tasks and various target models.}
To better verify the effectiveness of the proposed dynamic substitute structure learning strategy, we choose several basic deep learning networks, \emph{i.e.}, ResNet-18, ResNet-34, ResNet-50, and ResNet-101, as the substitute models. For a fair comparison, compared to `+ DSSL (Ours)' the basic substitute model also uses the graph-based structural information learning strategy. As shown in Tab.~\ref{Tab: Comparsion-DSSL}, we can draw the following conclusions,
(1) By applying the dynamic substitute structure learning strategy, our method can achieve much better attack results over all tasks and target models, such as comparing the `ResNet-34' with the next lower raw `+ DSSL (Ours)'. It is reasonable for different tasks and target models to have various corresponding structures as the best substitute models. Thus, our DSSL component adaptively generating more reasonable substitute structure can indeed promote the attack performance.
(2) For the same target model over the same dataset, the ASRs of `+ DSSL (Ours)' have little differences among various basic substitute model architectures. For example, to attack the AlextNet model trained on MNIST, our ASRs are around 69\% with small fluctuations based on four basic models under non-target attack setting. However, without applying the dynamic substitute structure learning strategy, the ASRs change dramatically from 47\% to 59\% according to the different substitute models. These results reinforce that our method can automatically produce optimal substitute structures over different basic models. 
(3) We also report the skip rate in the experiments, it is obvious that the number of skipped blocks is more when the basic substitute structure is complex with deeper structure, \emph{e.g.}, the skip rate is usually larger for utilizing ResNet-101 as basic substitute model compared to the ResNet-18. And the skip rate is usually smaller when targeting more difficult task, \emph{e.g.}, CIFAR-100 dataset, due to the substitute model keeping more blocks to learn more complex feature representations.


\begin{table}[t]
\begin{centering}
\hspace{-0.1in}
\setlength{\tabcolsep}{1.8mm}{
\begin{tabular}{c|cccc}
\hline
Method & ASRs & Distance & Train-Q & Test-Q \tabularnewline
\hline
GLS \cite{narodytska2016simple} & 53.28 & 3.68 & - & 311 \tabularnewline
Boundary \cite{brendel2017decision} & 99.32 & 3.94 & - & 702 \tabularnewline
\hline
DaST \cite{zhou2020dast} & 92.48 & 3.79 & 20M & - \tabularnewline
MAZE \cite{kariyappa2021maze} & 93.95 & 3.90 & 30M & - \tabularnewline
DDG+AST \cite{wang2021delving} & 93.29 & 3.88 & 15M & - \tabularnewline
\hline
Baseline-I  & 55.92 & 3.82 & 30M & - \tabularnewline
Baseline-II  & 70.28 & 3.91 & 30M & - \tabularnewline
+ GSIL  & 83.24 & 3.76 & 13M & - \tabularnewline
+ DSSL  & 95.02 & 3.80 & 9M & - \tabularnewline
\hline
\end{tabular}}
\caption{Comparison results among our DST method, variants of DST, and several competitors, targeting the Microsoft Azure example model and using BIM as the default non-target attack method. The `ASRs' refers to the attack success rate, `Distance' means the average perturbation distance per image, `Train-Q' is the number of queries in the model learning stage, and `Test-Q' denotes the query times during the evaluation. 
\label{Tab: Comparsion-Query}}
\par\end{centering}
\end{table}

\noindent \textbf{The efficacy of each proposed component for reducing the number of query times during the substitute training.}
Considering that in real-world applications, many online platforms defend themselves by detecting the number of query times for a single IP address, thus, more and more attack methods pay more attention to decreasing the number of query times. As illustrated in Tab.~\ref{Tab: Comparsion-Query}, we compare the number of query times during both the model learning and inference stage with competitors. 
(1) With a similar perturbation distance, our DST attack method not only achieves the comparable ASRs against others, but also needs the least query times during training and zero query times during attacking. 
As for the compared score-based and decision-based attacks, they need to query the target model to generate each attack in the evaluation stage, and feed the target model with the same original data numerous times which can be tracked easily by the defender.
(2) With the structural information learning strategy, the number of training query times drops significantly, which states that the graph-based relationship among the outputs can help $\mathcal{S}$ learn knowledge much quicker from the target one.
(3) With the proposed dynamic substitute structure learning strategy, the number of training query times of our attack method declines to some extent, which explain that $\mathcal{S}$ with optimal structure not only learn better but also learn faster from the $\mathcal{T}$.

\begin{figure}[t]
\begin{centering}
\hspace{-0.1in}
\includegraphics[scale=0.27]{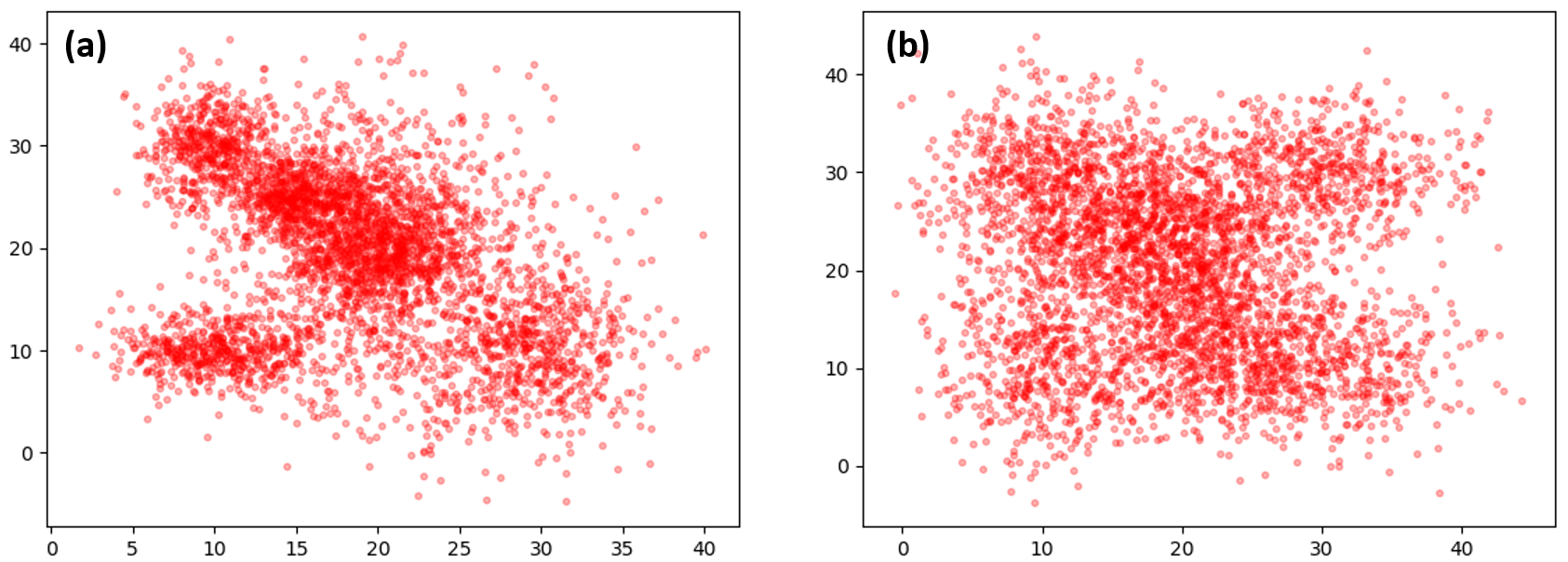} 
\par\end{centering}
\caption{Visualization of 3,000 generated data distribution using t-SNE \cite{maaten2008visualizing} on CIFAR-10. (a) Data generated by the $\mathcal{G}$ of DST without applying graph-based structural information learning strategy. (b) Data generated by the $\mathcal{G}$ of our DST. 
\label{fig:tsne}}
\end{figure}

\noindent \textbf{The efficacy of graph-based structural information learning strategy for the quality of generated training data.}
In terms of feature, we visualize the feature distribution of synthesized data extracted by the target model in Fig.~\ref{fig:tsne}. Such qualitative results show that, with applying the proposed graph-based structural information learning strategy, the generated data is more evenly distributed, which is friendly for substitute model training.

\begin{figure}[t]
\begin{centering}
\hspace{-0.2in}
\includegraphics[scale=0.19]{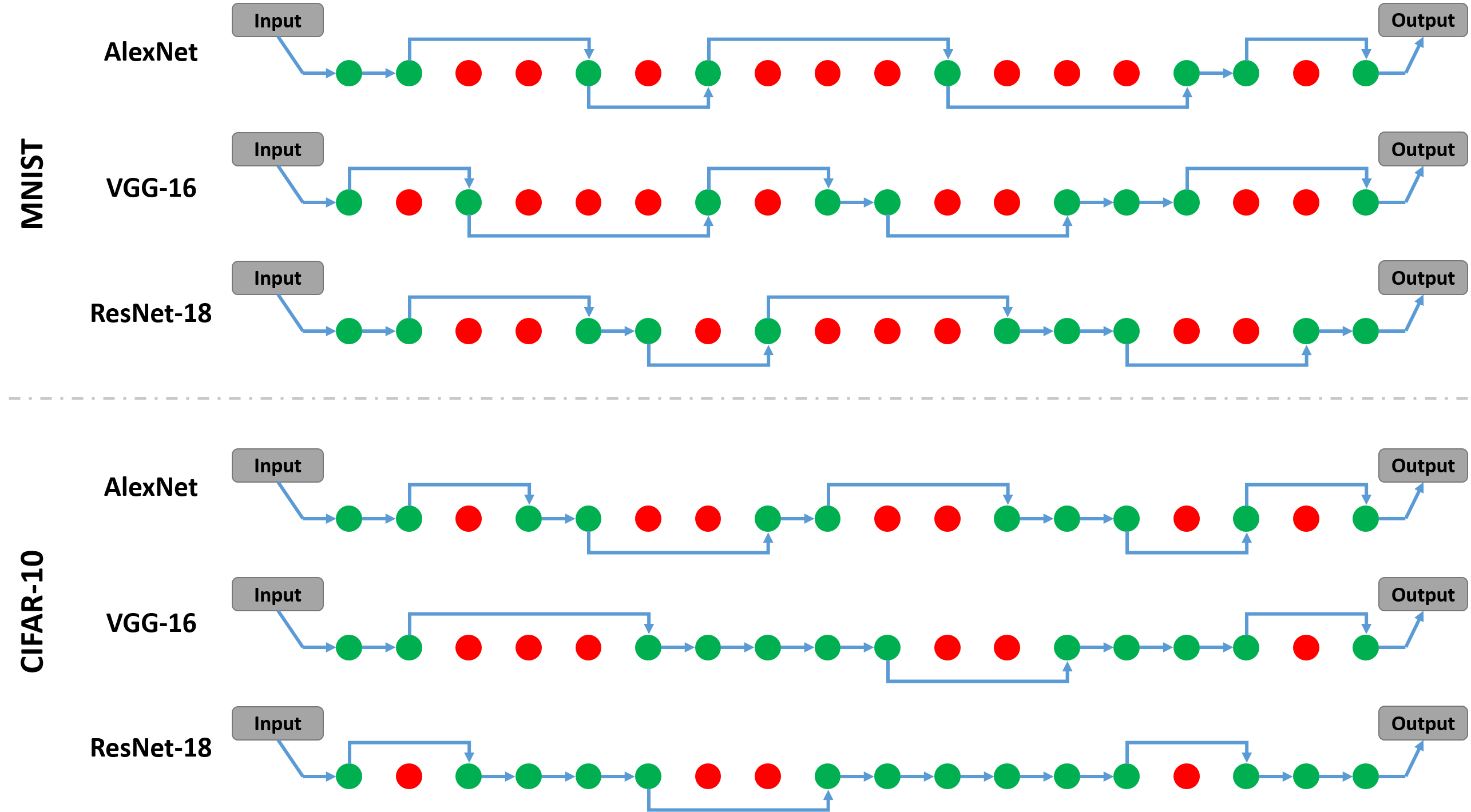} 
\par\end{centering}
\caption{The visualizations of the dynamic substitute model structures according to different target models (\emph{i.e.}, AlexNet, VGG-16, and ResNet-18) and tasks (\emph{i.e.}, MNIST and CIFAR-10). The basic network of substitute models is ResNet-18. The green dot means the keeping block, the red dot represents the skipped block.
\label{fig:skipnet}}
\end{figure}

\noindent \textbf{The visualizations of dynamic substitute model structures according to different target models and tasks.}
As shown in Fig.~\ref{fig:skipnet}, for the same basic substitute model architecture, the optimal substitute models, generated by our dynamic substitute structure learning strategy, are different according to various targets. Considering that the MNIST dataset is relatively simple compared to CIFAR-10, the number of skipped blocks is more and keeping blocks is fewer for attacking the MNIST model. Meanwhile, the AlexNet expresses relatively simple representations, thus, the corresponding optimal substitute models keep fewer blocks to learn knowledge from AlexNet.

\noindent \textbf{Social impact and limitations.}
The task of adversarial attack is predominantly utilized as a tool for the verification and validation the robustness of the state-of-the-art deep models. Our DST method shall not be utilized to attack existing recognition systems.  
On the other hand, our DST still relies on white-box attack methods, we will focus on attacking directly and study the defense algorithm in future.

\section{Conclusion}
To tackle the data-free black-box attack task, we propose a novel dynamic substitute training attack method (DST). DST generate the optimal substitute model structure according to the different targets by the proposed dynamic substitute structure learning strategy, and encourage the substitute model to learn implicit information from the target one via graph-based structure information learning constrain. The experiments show that DST can achieve the best attack performance compared with existing methods.

\section{Acknowledgement}
This work was supported in part by NSFC Project (62176061), Shanghai Municipal Science and Technology Major Project (2018SHZDZX01), and Shanghai Research and Innovation Functional Program (17DZ2260900).

\clearpage
{\small
\bibliographystyle{ieee_fullname}
\bibliography{egbib}
}

\end{document}